\documentclass{article} 
\usepackage{iclr2025_conference,times}
\usepackage{graphicx}
\usepackage[subrefformat=parens]{subcaption}
\usepackage{booktabs}

\usepackage{amsmath,amsfonts,bm}

\def\eqref#1{equation~\ref{#1}}

\def\1{\bm{1}}

\DeclareMathAlphabet{\mathsfit}{\encodingdefault}{\sfdefault}{m}{sl}
\SetMathAlphabet{\mathsfit}{bold}{\encodingdefault}{\sfdefault}{bx}{n}

\usepackage{hyperref}
\usepackage{url}

\title{Accelerating High-Efficiency Organic Photovoltaic Discovery via Pretrained Graph Neural Networks and Generative Reinforcement Learning}

\author{
  \textbf{Jiangjie Qiu}$^{1}$ \ \ \ \textbf{Hou Hei Lam}$^2$ \ \ \ \textbf{Xiuyuan Hu}$^{3}$ \ \ \ \textbf{Wentao Li}$^{1}$ \ \ \  \textbf{Siwei Fu}$^{1}$\ \ \
  \\ \textbf{Fankun Zeng}$^{4}$\ \ \ \textbf{Hao Zhang}$^{3}$\ \ \ \textbf{Xiaonan Wang}$^{1}$\thanks{Corresponding author, email: wangxiaonan@tsinghua.edu.cn}\\ 
  $^1$Department of Chemical Engineering, Tsinghua University\\
  $^2$Department of Automation, Tsinghua University \\
  $^3$Department of Electronic Engineering, Tsinghua University\\
  $^4$Weiyang College, Tsinghua University\\ 
}
\iclrfinalcopy 
\begin{document}

\maketitle

\begin{abstract}
Organic photovoltaic (OPV) materials offer a promising avenue toward cost-effective solar energy utilization. However, optimizing donor-acceptor (D-A) combinations to achieve high power conversion efficiency (PCE) remains a significant challenge. In this work, we propose a framework that integrates large-scale pretraining of graph neural networks (GNNs) with a GPT-2 (Generative Pretrained Transformer 2)-based reinforcement learning (RL) strategy to design OPV molecules with potentially high PCE. This approach produces candidate molecules with predicted efficiencies approaching 21\%, although further experimental validation is required. Moreover, we conducted a preliminary fragment-level analysis to identify structural motifs recognized by the RL model that may contribute to enhanced PCE, thus providing design guidelines for the broader research community. To facilitate continued discovery, we are building the largest open-source OPV dataset to date, expected to include nearly 3,000 donor-acceptor pairs. Finally, we discuss plans to collaborate with experimental teams on synthesizing and characterizing AI-designed molecules, which will provide new data to refine and improve our predictive and generative models.
\end{abstract}

\section{Introduction}
Organic photovoltaics (OPVs) have attracted broad attention in the research community because of their potential to enable lightweight, flexible, and cost-effective solar cells. One of the primary challenges in advancing OPV technology lies in identifying donor-acceptor pairs with outstanding power conversion efficiencies (PCEs). In the past, designing new donors and acceptors relied heavily on labor-intensive trial-and-error experiments and incremental modifications to existing structures. Machine learning (ML) and data-driven approaches can substantially reduce the time and cost of this design process by efficiently exploring large chemical design spaces and providing rapid performance predictions. However, the success of ML-based molecular design depends on the availability of high-quality datasets and on models capable of capturing subtle structure-property relationships. A review of the relevant literature indicates that most prior studies have focused on either a single system \citep{Sun2024} or on property prediction alone. In the few instances where molecular generation or exploration is addressed, methods such as screening and combining existing molecular structures, variational autoencoders (VAEs), or genetic algorithms are employed. \citep{Hutchison2023}\citep{Sun2024}\citep{ZhangSA2025}, these approaches typically optimize known structures, exhibiting relatively limited capacity to generate and explore genuinely novel molecules. Furthermore, the majority of these studies employ donor-acceptor datasets of about 500-1,500 entries \citep{Saeki2021}\citep{Min2020}\citep{Hutchison2023}, covering relatively narrow material systems and thereby limiting generalizability to out-of-distribution structures.

In this work, we combine large-scale graph neural network (GNN) pretraining with reinforcement learning (RL) to optimize donor-acceptor pairs for higher PCE. By integrating quantum-level property prediction tasks (specifically, highest occupied molecular orbital (HOMO) and lowest unoccupied molecular orbital (LUMO) energies) with a molecular masking and reconstruction task, we obtain embeddings that effectively capture multi-level information. We then apply self-attention and cross-attention modules to fuse the donor and acceptor embeddings, resulting in more accurate PCE predictions than those produced by methods relying on molecular fingerprints or simple GNN-based embeddings. These predictions subsequently serve as a reward function within a GPT-2-based RL strategy to generate new candidate molecules. Through multiple rounds of optimization, the model converges on structures with progressively higher predicted PCE, thereby demonstrating the feasibility of our end-to-end, AI-driven design pipeline. Concurrently, we are finalizing the release of the largest curated OPV dataset to date, which is expected to include nearly 3,000 donor-acceptor pairs, with the aim of accelerating further innovation in the OPV research community.

\section{Methodology}
Our research approach begins with constructing effective embeddings for donor and acceptor molecules. To build robust embeddings, we pretrain a GNN using approximately 51k organic small molecules \citep{Lopez201751k}, each associated with SMILES (Simplified Molecular Input Line Entry System) notations\citep{smiles} and corresponding HOMO and LUMO data.

The pretraining process primarily involves two core tasks. The first task is molecular masking and reconstruction, similar to the Masked Language Model (MLM) in language models. In this process, SMILES strings are converted into graph representations, and some structures are masked to encourage the model to recover the missing elements. This strategy helps the GNN learn meaningful chemical features and connectivity patterns. The second task uses the aforementioned data to predict HOMO and LUMO energies, enabling the model to directly capture the electronic properties that are vital for OPV performance. Once pretraining is complete, the resulting embeddings are used for donor-acceptor property prediction. We then incorporate self-attention and cross-attention mechanisms to capture interactions between donor and acceptor embeddings, thereby highlighting the significance of specific functional groups and structural motifs. Finally, the fused feature representations are fed into a multilayer perceptron (MLP) to output predicted PCE.

\begin{figure}[h]
\begin{center}
    \includegraphics[width=0.75\textwidth]{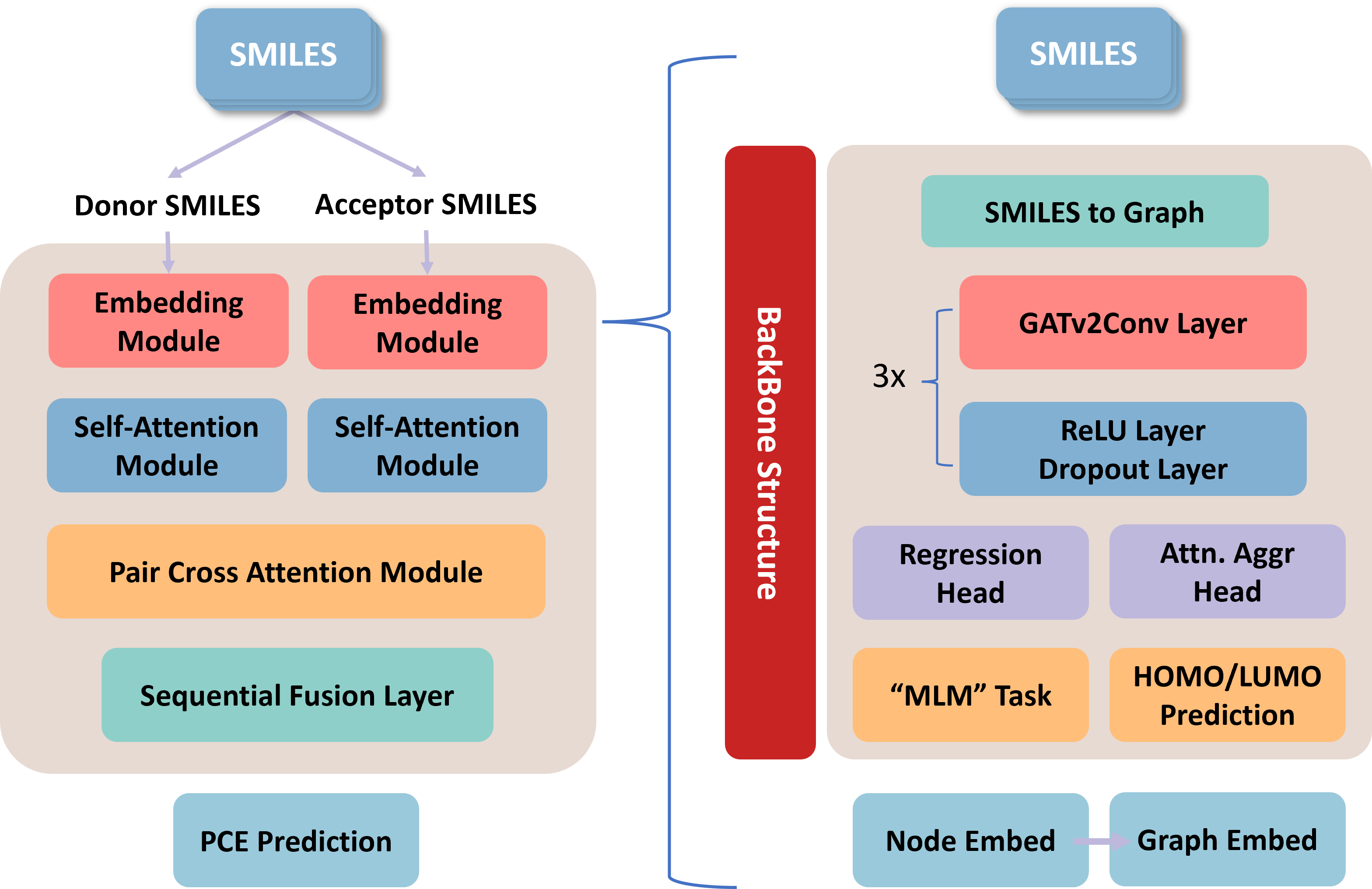} 
\end{center}
\caption{The Structure of the Pretrain Model and Predictor Model.}
\end{figure}

To explore novel molecular designs, we introduce a generator based on the classical autoregressive language model architecture, GPT-2 \citep{GPT2}, into the reinforcement learning loop \citep{Hu2024RL}. In this process, a partially specified molecule (either a donor or an acceptor) serves as a prompt, and the generator produces candidate structures for the missing counterpart. The pre-trained power conversion efficiency predictor acts as a reward function, guiding the generator to generate more efficient molecule pairs in successive iterations. Chemical screening methods, such as basic rule checks and RDKit validations, ensure that the generated molecules are structurally valid. After multiple rounds of optimization, the generator will converge to a set of donor-acceptor (D-A) structures with high predicted PCE values. Although these values need to be experimentally verified, they highlight the potential of combining graph neural network-based property prediction with reinforcement-learning-driven generation for the discovery of organic photovoltaics.

\begin{figure}[h]
    \centering
    \begin{subfigure}[b]{0.27\textwidth}
        \includegraphics[width=\textwidth]{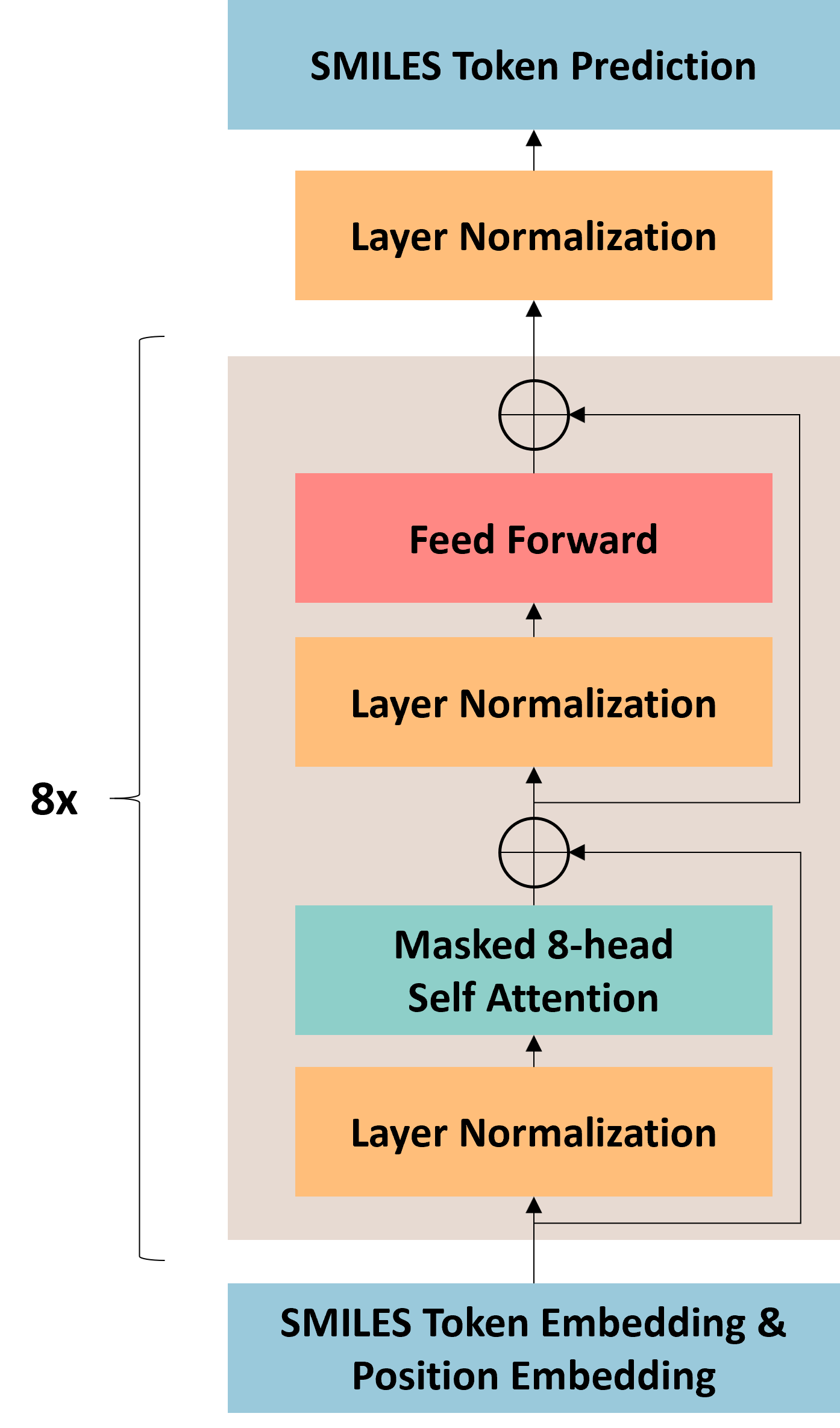}
        \caption{}
        \label{fig:subfig_a}
    \end{subfigure}
    \hfill
    \begin{subfigure}[b]{0.7\textwidth}
        \includegraphics[width=\textwidth]{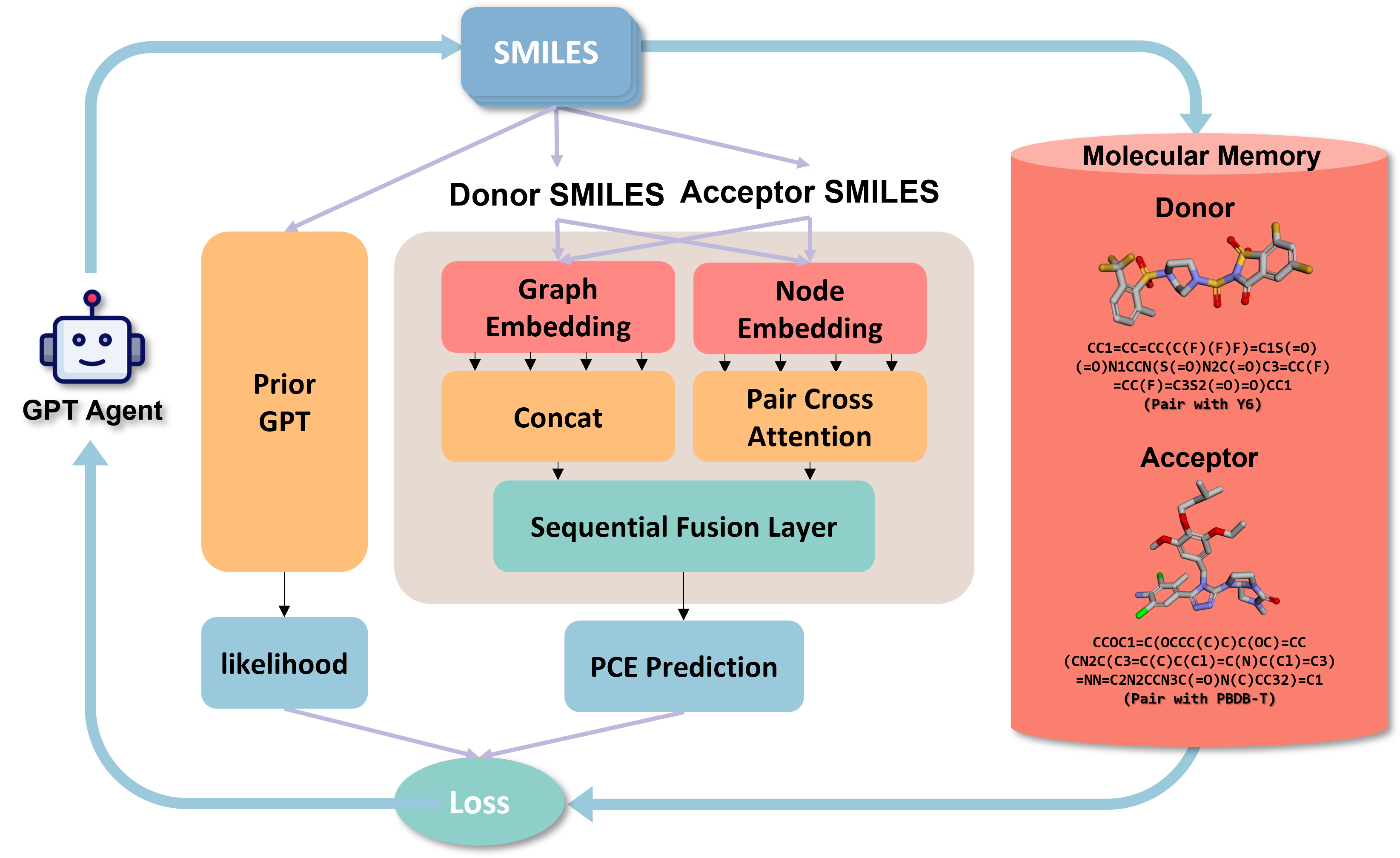}
        \caption{}
        \label{fig:subfig_b}
    \end{subfigure}
    \caption{(a) The architecture of the GPT. (b) The Structure of the RL model.}
    \label{fig:fig_rl}
\end{figure}

The proposed GPT architecture, as illustrated in Fig.\ref{fig:fig_rl}(a), comprises a hierarchical Transformer-based generator with three core components: 1) Embedding Layer combining SMILES token embeddings and learnable positional encodings; 2) Stacked Transformer Blocks featuring masked 8-head self-attention with causal constraints, layer normalization, and feed-forward networks with quadruple hidden dimension; 3) Prediction Head generating token probabilities through final layer normalization and linear projection. The proposed reinforcement learning framework integrates policy optimization with knowledge preservation through a dual-objective loss formulation. Each GPT agent, initialized from a shared pre-trained generative prior, optimizes a squared error function combining reward maximization and policy regularization as \begin{equation}
\mathcal{L}(x;\theta) = \left[ \log P_{\text{Prior}}(x) - \log P_{\text{Agent}}(x;\theta) + \sigma \cdot s(x) \right]^2
\label{eq:loss_function }
\end{equation}
where $x$ is the generated molecular SMILES string, $\log P_{\text{Prior}}$ denotes the pretrained generator's log-likelihood, $\log P_{\text{Agent}}$ represents the learnable agent distribution parameterized by model weights $\theta$. This design introduces three essential components: 1) The divergence term $\log P_{\text{Prior}} - \log P_{\text{Agent}}$ enforces proximity to the original molecular generation distribution, effectively preventing catastrophic forgetting through implicit KL divergence regularization; 2) The reward term 
$s(x)$ injects PCE preference signals via learned scoring functions; 3) The quadratic formulation automatically balances magnitude discrepancies between policy constraints and reward signals. The reward coefficient $\sigma$ dynamically decays based on achieved score improvements to stabilize policy optimization. By simultaneously optimizing syntactic validity through likelihood preservation and chemical desirability through score maximization, this loss architecture enables stable policy improvement while maintaining fundamental molecular syntax constraints inherited from the pre-trained generator.

\section{Experiments and Results}
We have curated a dataset of approximately 2,500 donor-acceptor pairs, each featuring experimentally measured PCE values and various related properties (such as $J_{sc}$, $V_{oc}$, fill factor, and donor/acceptor band energies). This dataset spans a diverse range of polymer and small-molecule donors, as well as both fullerene and non-fullerene acceptors. We compared our pretrained model with an unpretrained model and a baseline model built on molecular fingerprints and random forests. The results show that knowledge gained from the molecular masking-reconstruction and HOMO/LUMO prediction tasks substantially reduces the mean squared error (MSE) in PCE prediction.

During the reinforcement learning phase, our model undergoes iterative updates guided by a reward signal based on power conversion efficiency (PCE)(see Fig.\ref{fig:1_pce}). A chemical validity screening mechanism ensures that the generated molecules are synthetically feasible. To date, we have designed corresponding donor and acceptor molecules for the widely studied acceptor Y6 \citep{Y6} and donor PBDB-TF \citep{PM6},(see Fig.\ref{fig:PBDB-TF}, \ref{fig:Y6}). Notably, among the newly generated donor molecules designed for Y6, the top-ranked candidates exhibit predicted PCE values exceeding 21\%, whereas the highest PCE recorded in the dataset for Y6-based systems is approximately 19\%. Although experimental validation is still required, these findings underscore the significant enhancement in virtual screening capabilities.

We also performed a preliminary fragment-level frequency analysis on the generated donors and acceptors. The analysis revealed recurring aromatic scaffolds and electron-deficient moieties that the model appears to favor. For donor structures, extended conjugation motifs and electron-rich cores were particularly common, while in acceptor structures we frequently observed strongly electron-withdrawing groups, which may contribute to enhanced charge-transfer stability. Frequent halogenation and specific substitution patterns suggest that the model has learned key design principles for balancing energy levels and improving blend morphology. Nonetheless, our fragment-level analysis remains at an early stage, and further in-depth exploration incorporating expert knowledge is necessary.

\section{Conclusion and Future Work}
We propose a unified approach that integrates graph neural network (GNN) pretraining and reinforcement learning to advance the design of organic photovoltaic (OPV) materials. A reinforcement learning-based generator, built upon the GPT-2 architecture, generates donor and acceptor candidates with predicted power conversion efficiencies potentially exceeding 20\%. Equally important, we will conduct a detailed molecular fragment analysis to elucidate the molecular features underlying these high-efficiency predictions, providing valuable insights for further optimization and experimental validation.

Looking ahead, we plan to collaborate with experimental laboratories to synthesize and test the most promising candidate materials identified by our reinforcement learning framework. The results from these experiments will be incorporated into our model, facilitating continuous improvements in both the performance predictor and the molecular generator. Additionally, we are preparing to release the largest curated OPV dataset to date, expected to contain nearly 3,000 donor-acceptor pairs, with the aim of accelerating innovation in the OPV research community. We believe that our approach, which integrates data-driven embedding representations, attention-based interaction modeling, and reinforcement learning, represents a significant step toward more efficient and targeted discovery of high-performance organic photovoltaic materials.

\section{Acknowledgement}
We acknowledge support from Tsinghua University Initiative Scientific Research Program (Student Academic Research Advancement Program: Zhuiguang Special Project) (Grant No. 20247020006)
\label{others}

\bibliography{iclr2025_conference}
\bibliographystyle{iclr2025_conference}

\appendix

\section{Appendix}

\subsection{Dataset Details}

\paragraph{Organic Small Molecule Dataset:}  
The dataset used for GNN pretraining consists of approximately 51,000 organic small molecules, sourced from the dataset published by \citet{Lopez201751k}. The molecules were constructed by assembling molecular functional fragments, and their HOMO and LUMO energy levels were obtained via DFT calculations and subsequently calibrated using experimental data.

\begin{figure}[h]
\begin{center}
    \includegraphics[width=0.75\textwidth]{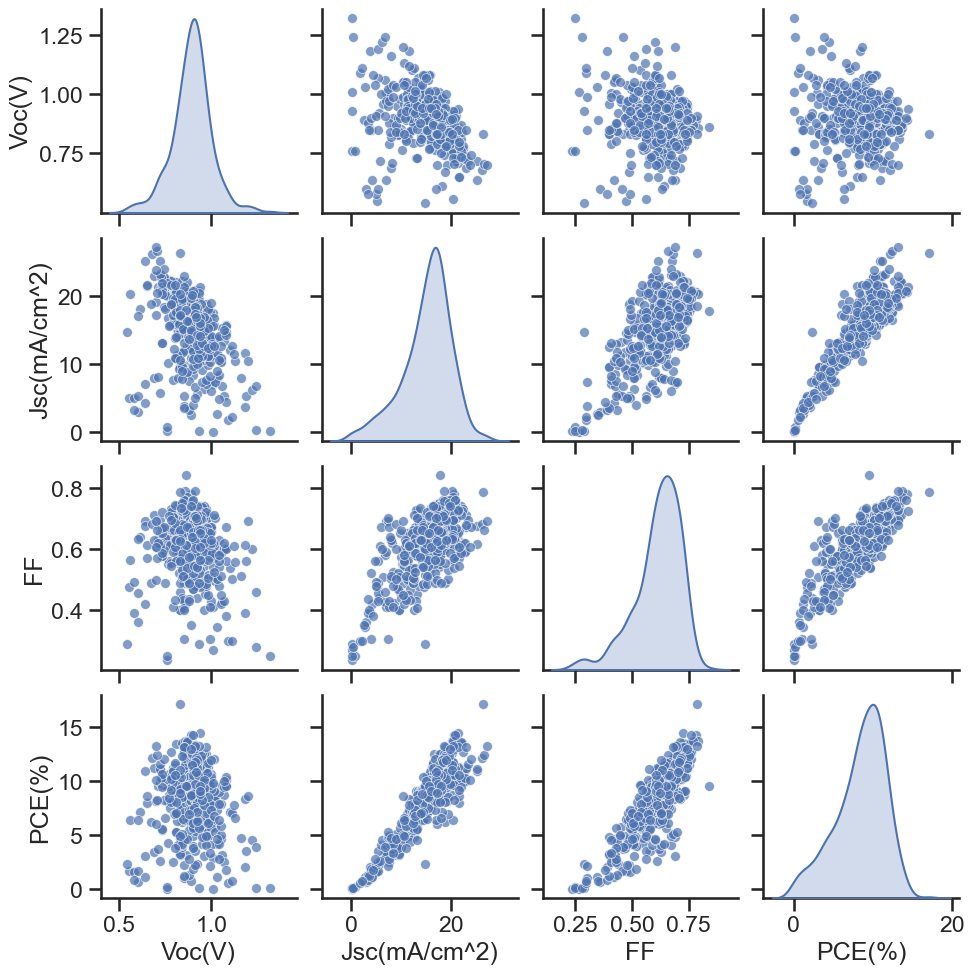} 
\end{center}
\caption{The distribution of device parameters}\label{fig:data}
\end{figure}

\paragraph{Donor-Acceptor Pair Dataset:}  
This dataset currently contains approximately 2,500 donor-acceptor pairs, with an expected final dataset size of around 3,000 pairs. The distribution of device parameters of the current datasets can be seen in Fig.\ref{fig:data}. During dataset construction, we first collected open-source datasets published by previous researchers \citep{Hutchison2023}\citep{Saeki2021}\citep{Min2020}, performed deduplication and calibration, and corrected potential errors in SMILES representations that may have resulted from recording mistakes. Additionally, we standardized the recording format for various parameters.

Since data collected by different researchers may follow distinct conventions—such as truncating molecular side chains to ethyl groups or simplifying polymers to trimers or hexamers \citep{Hutchison2023}, data from multiple sources could not be directly merged. To address this issue, we manually verified each dataset entry and re-recorded the data using a unified methodology. We are actively collecting additional device data, and each dataset entry includes precise records of power conversion efficiency (PCE) and related properties, such as (\(\mathrm{J_{sc}}\), \(\mathrm{V_{oc}}\), \(\mathrm{FF}\), and energy levels), along with detailed information on data sources.

\subsection{Model Hyperparameter Settings}
\paragraph{GNN Model:}
The number of layers in the Graph Neural Network (GNN) is set to 3. Each layer consists of a GATv2Conv layer \citep{GATv2}, a non-linear layer, and a Dropout layer. In the molecular masked reconstruction task, the masking ratio is set to 21\%. For the prediction of the Highest Occupied Molecular Orbital (HOMO) / Lowest Unoccupied Molecular Orbital (LUMO), AttentionalAggregation is used for attention pooling, and the final results are regressed. For both tasks, AdamW is used as the optimizer, and the learning rate is dynamically adjusted using ReduceLROnPlateau.

\paragraph{PCE Predictor Model:} The model employs a dual-encoder architecture in which two pre-trained graph neural networks independently extract both graph-level and node-level embeddings for donor and acceptor molecules. Bidirectional cross-attention is then applied to the node-level embeddings using the Multihead Attention module with 8 heads and dropout, enabling donor-to-acceptor and acceptor-to-donor interactions. The resulting attention outputs are aggregated via mean pooling and subsequently concatenated with the corresponding graph-level embeddings to form a fused representation. This fused feature vector is processed by a regression head consisting of a linear layer, a ReLU activation, a dropout, and a final linear layer to generate the prediction. AdamW is utilized as the optimizer, with the learning rate dynamically adjusted through the ReduceLROnPlateau scheduler.

\paragraph{GPT-2-Based Generator:}
The fine-tuning phase employs 1000 training iterations with a batch size of 128. The reinforcement learning framework implements a linear reward modulation strategy where the reward coefficient ($\sigma = 100$) decays proportionally to training progress, without explicit reward discount factors. Molecular exploration integrates multinomial sampling-based probabilistic generation of SMILES with top-5\% experience replay from episodic memory.

\subsection{Supplementary Details on Molecular Design}
\begin{figure}[h]
\begin{center}
    \includegraphics[width=0.7\textwidth]{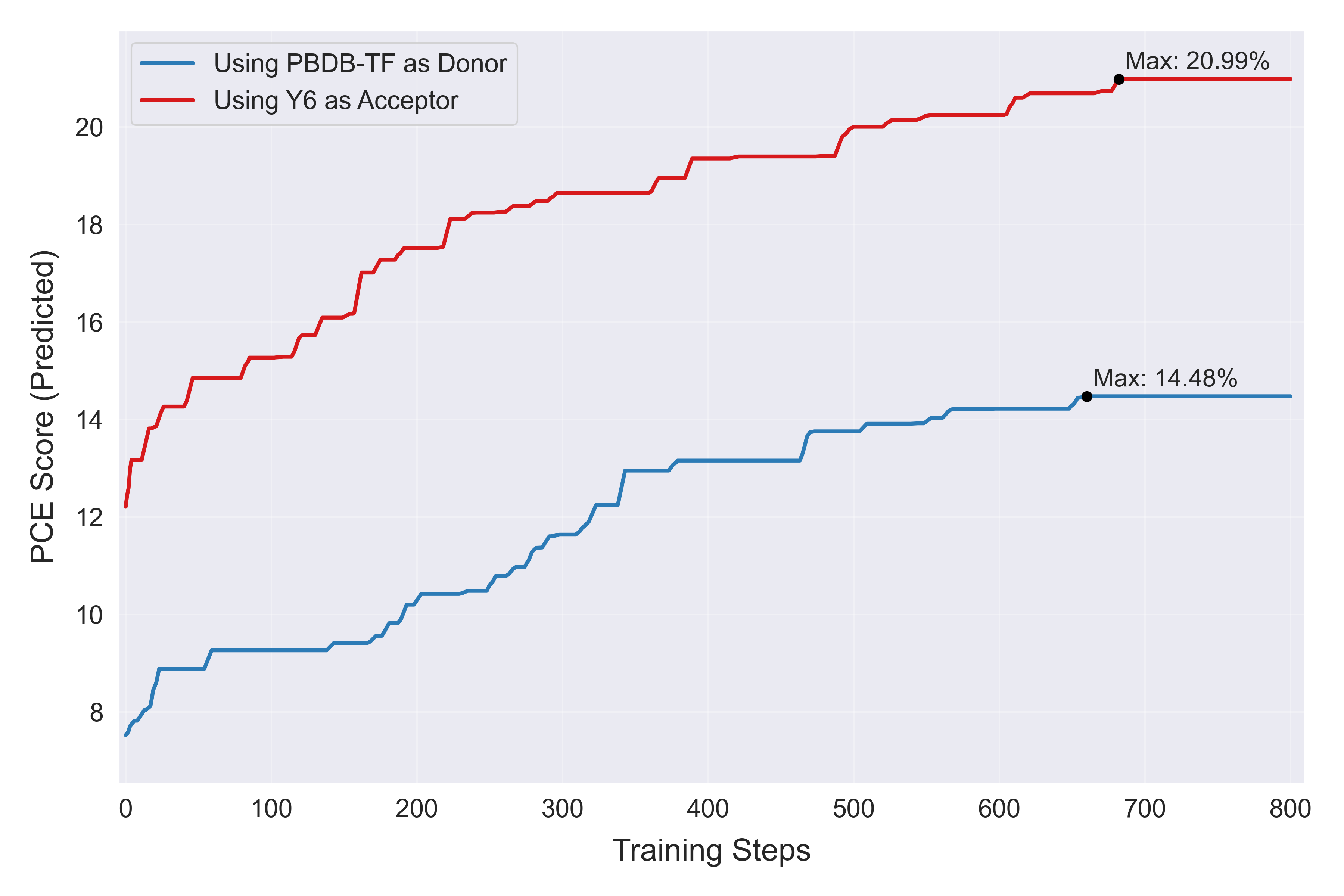} 
\end{center}
\caption{Top-1 PCE Score Trend for Generated Donor-Acceptor Pairs}\label{fig:1_pce}
\end{figure}

Fig.\ref{fig:1_pce} illustrates the evolution of the top-1 PCE values derived from the donor-acceptor pairs stored in memory as the GPT-2-Based Generator undergoes fine-tuning. Notably, the trend indicates a progressive enhancement in the generated pairs' performance, culminating in a maximal PCE value that underscores the model's optimization capabilities. In particular, at training step 800, when using PBDB-TF as the donor, the maximum observed PCE reaches 14.48\%, whereas employing Y6 as the acceptor yields a maximum of 20.99\%.

\begin{figure}[htbp] 
    \centering 
    \includegraphics[width=0.8\textwidth]{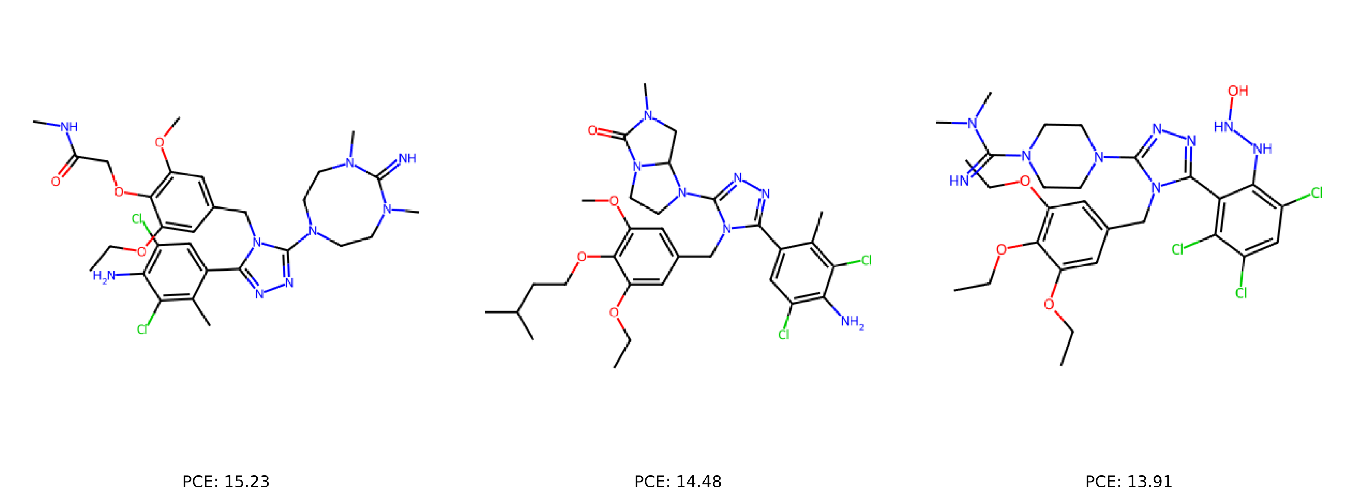} 
    \caption{Some of the Top Acceptors designed for PBDB-TF} 
    \label{fig:PBDB-TF} 

    \centering
    \includegraphics[width=0.8\textwidth]{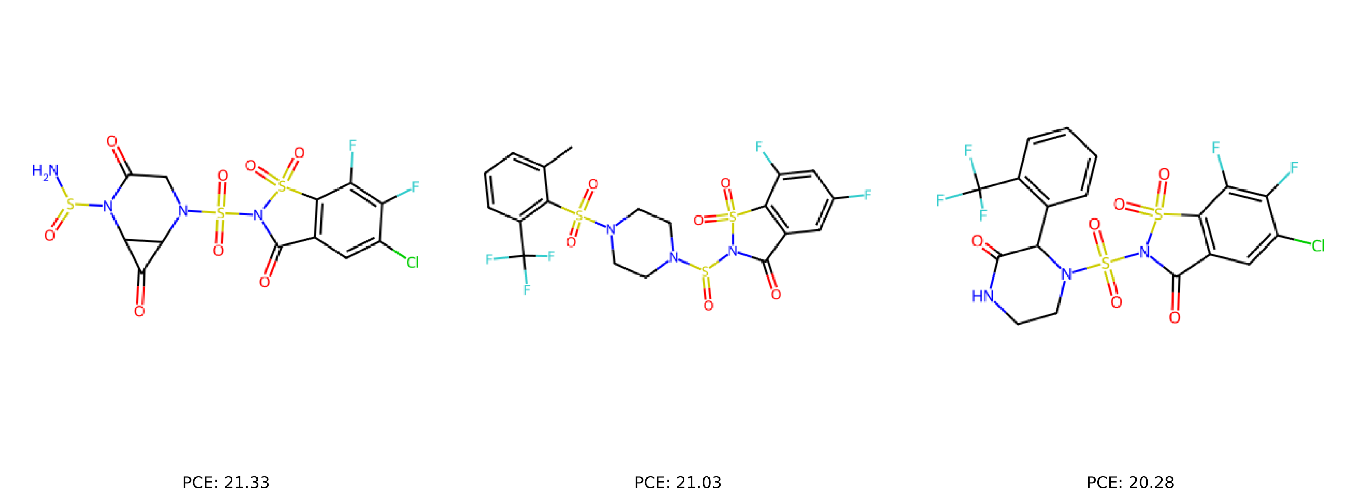}
    \caption{Some of the Top Donors designed for Y6}
    \label{fig:Y6}
\end{figure}

\end{document}